\documentclass[letterpaper]{article}
\usepackage[preprint]{aaai2027}  
\usepackage[hyphens]{url}  
\usepackage{graphicx} 
\usepackage{natbib}  
\usepackage{caption}
\usepackage{algorithm}
\usepackage{algorithmic}

\usepackage{newfloat}
\usepackage{listings}
\DeclareCaptionStyle{ruled}{labelfont=normalfont,labelsep=colon,strut=off} 
\floatstyle{ruled}
\newfloat{listing}{tb}{lst}{}
\floatname{listing}{Listing}

\usepackage{booktabs}

\title{Beyond the Text: Verifying That Agent-Written Papers Are Backed by Their Artifacts}
\author{
Qiuhong Shen\textsuperscript{\rm 1},
Benlong Wu\textsuperscript{\rm 1},
Hanjin Liu\textsuperscript{\rm 1},
Yuang Qi\textsuperscript{\rm 1},
Kejiang Chen\textsuperscript{\rm 1}\corresponding
}

\affiliations{
\textsuperscript{\rm 1}University of Science and Technology of China
}

\begin{document}

\maketitle

\begin{abstract}
Large language model agents are increasingly capable of conducting research autonomously, producing research documents alongside the code and experiments that ostensibly support them. Yet whether the reported findings are consistently supported by corresponding implementations and execution evidence remains largely unexplored: existing review practices primarily assess textual quality and cannot reliably identify inconsistencies such as hard-coded metrics, unimplemented methods, or unsupported experimental results. We present ReAgent, an automated auditing framework for assessing the consistency between agent-generated research documents and their associated repositories. ReAgent constructs structured representations of scientific claims from research documents and uses them to guide repository analysis and evidence collection. Static auditing examines whether claimed methodologies, implementations, and experimental configurations are consistently reflected in the repository, while dynamic auditing executes relevant experiments and collects execution evidence to assess empirical findings. By combining static analysis with dynamic evidence, ReAgent identifies inconsistencies that may remain hidden under either perspective alone, such as experiments that reproduce reported numbers while deviating from the claimed methodology. The collected evidence and audit decisions are organized into a structured repository-level audit report, enabling transparent evidence traceability. We evaluate ReAgent on a manually curated benchmark of agent-generated research document--repository pairs and compare it against representative static and reproduction-based baselines. Experimental results demonstrate that ReAgent effectively identifies inconsistencies between reported research findings and their supporting repository evidence.
\end{abstract}

\section{Introduction}

In recent years, rapid advances in Large Language Models (LLMs) have significantly expanded the capabilities of intelligent agents~\cite{wang2024survey,jia2026autotool}. From software engineering agents~\cite{jimenez2024swe,yang2024swe} to embodied agents~\cite{liu2025open,lin2023grounded} and autonomous research agents~\cite{zheng2025automation,zamprogno2025autonomous}, these systems are evolving into autonomous agents capable of planning, executing, and refining complex workflows. In scientific research, recent systems have demonstrated the ability to perform literature review, hypothesis generation, code implementation, experiment execution, result analysis, and paper writing, covering increasingly large portions of the research pipeline. Representative systems such as the AI Scientist~\cite{lu2026towards}, VirSci~\cite{su2025many}, Agent Laboratory~\cite{schmidgall2025agent}, and FARS~\cite{tang2026fars} highlight the growing potential of autonomous research.

Existing studies primarily evaluate whether autonomous research agents can generate novel scientific ideas and high-quality research outputs~\cite{si2024can,chen2025mlrbench,wang2026act,chen2025scienceagentbench,tang2026ai}. As these systems increasingly automate end-to-end scientific workflows, a more fundamental yet largely overlooked question arises:

\begin{center}
\textit{Are the scientific claims reported by autonomous research agents trustworthy?}
\end{center}

This issue is not unique to autonomous research. In traditional scientific practice, inconsistencies between papers and their implementations have long contributed to the reproducibility crisis~\cite{baker20161,gundersen2018state,pineau2021improving}. Autonomous research further amplifies this challenge because research documents, code, and experiments are often generated and iteratively refined independently. Combined with well-known limitations of foundation models, including hallucination~\cite{ji2023survey,tian2025codehalu}, weak alignment between natural language and executable code~\cite{baumgartner2026scicoqa}, and limited self-verification~\cite{huang2024selfcorrect,tyen2024llms}, scientific claims may gradually diverge from their implementations and execution evidence. These characteristics introduce new challenges in maintaining consistency among scientific claims, code implementations, and experimental evidence in autonomous research pipelines, motivating the need for dedicated mechanisms to verify their alignment.

\begin{figure}[htbp]
    \centering
    \includegraphics[width=1\linewidth]{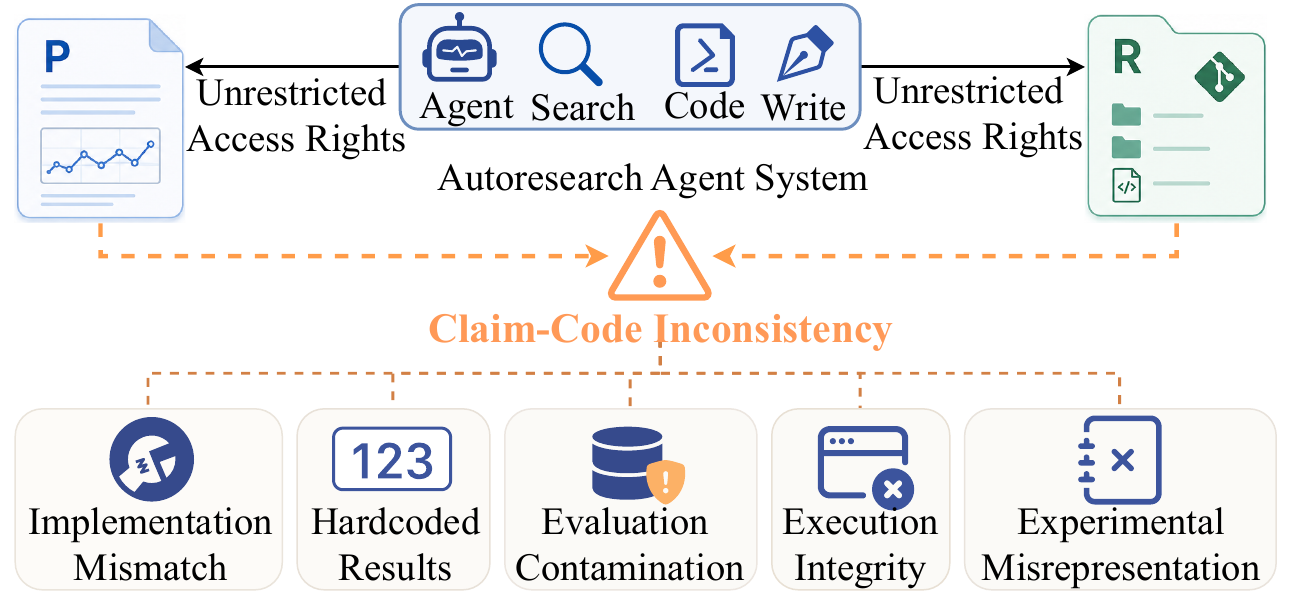} 
    \caption{Problem definition. An automated research agent generates a research document $P$ and an associated repository $R$. During autonomous generation and iterative refinement, claim-code inconsistencies may arise across the research workflow. The examples are illustrative rather than exhaustive; see Section ~\ref{sec:taxonomy} for the complete categorization.}
    \label{fig:Problem-Definition}
\end{figure}

Fundamentally, scientific research is trustworthy only when its claims are supported by empirical evidence~\cite{wadden2020fact,luo2025more}. Methodological claims, experimental protocols, and reported results should all be grounded in corresponding implementations, configurations, and execution evidence. However, discrepancies may arise when proposed methods are absent from the implementation, reported results cannot be reproduced, or experimental settings differ from those described in the paper. Although such inconsistencies do not necessarily imply malicious intent, they undermine the credibility of autonomous research outputs.

Regrettably, existing work has yet to establish a systematic evaluation framework targeting this issue. While recent studies on automated paper replication and autonomous research have advanced scientific automation, they primarily focus on code quality or peer-review scores~\cite{starace2025paperbench,chen2025deep,tang2026ai}, rather than whether scientific claims are supported by both implementation and experimental evidence. This challenge stems from the intrinsic dual nature of scientific claim verification: trustworthy assessment requires verifying both that the implementation matches the claimed methodology and that the reported results are supported by executable evidence. Static analysis can examine whether a repository contains implementations corresponding to described methods, but cannot determine whether reported findings were actually produced by the current implementation. Conversely, dynamic re-execution can verify whether results are reproducible, but cannot establish whether the executed implementation faithfully matches the scientific claims. Therefore, neither perspective alone is sufficient. Autonomous research requires a unified framework that jointly reasons over implementation consistency and executable evidence.

To bridge this gap, we introduce claim-code consistency auditing and ReAgent, an automated framework that systematically examines agent-generated research documents and repositories through complementary static verification and dynamic verification. The source code will be publicly available at \url{https://github.com/hongsq12345/ReAgent}. 

The main contributions of this paper are summarized as follows:
\begin{itemize}
    \item We introduce \textbf{Claim-Code Consistency} as a new auditing problem for autonomous research, defining how scientific claims should be verified against both code implementations and executable evidence.
    
    \item We propose \textbf{ReAgent}, an automated claim verification framework that combines static implementation auditing with dynamic evidence collection to systematically verify research documents and repositories generated by autonomous research agents.
    
    \item We establish a benchmark of agent-generated research repositories, enabling systematic evaluation of claim-code consistency and providing empirical characterization of inconsistency patterns in autonomous research outputs.
\end{itemize}
\section{Related Work}

\noindent\textbf{Autonomous Research Agents.}
Recent advances in large language models have enabled autonomous research agents capable of performing increasingly complete scientific workflows. Representative systems, including The AI Scientist~\cite{lu2026towards}, Agent Laboratory~\cite{schmidgall2025agent}, AI-Researcher~\cite{tang2026ai}, VirSci~\cite{su2025many}, AutoSOTA~\cite{li2026autosota}, ScientistOne~\cite{meng2026scientistone}, and AutoResearchClaw~\cite{liu2026autoresearchclaw}, demonstrate increasingly capable research automation spanning literature review, idea generation, implementation, experimentation, and paper writing. Existing work primarily evaluates research capability, such as workflow completion, scientific quality, or research productivity, while paying limited attention to whether reported scientific claims are faithfully supported by implementations and experimental evidence. Our work complements this line of research by treating generated research outputs and repositories as auditable objects and explicitly verifying their consistency.

\noindent\textbf{Trustworthy LLM Agents.}
The increasing autonomy of LLM agents has raised growing concerns regarding their trustworthiness. Recent studies investigate undesirable behaviors such as deceptive reasoning, specification gaming, and hidden coordination~\cite{huang2026deceptionbench, DBLP:conf/ijcnlp/MathewMMVWCS25}. In the scientific domain, prior work has revealed hidden implementation pitfalls in AI scientist systems~\cite{luo2025more} and evaluated whether frontier LLM agents can reliably conduct realistic research activities. These studies mainly focus on agent behavior during task execution, whereas we investigate whether the final research outputs faithfully support the scientific claims they report. 

\noindent\textbf{AI-assisted Scientific Auditing and Verification.}
Recent work has employed LLM agents to reconstruct implementations from papers or automatically reproduce reported experimental results, including PaperBench~\cite{starace2025paperbench}, DiscoveryBench~\cite{majumder2024discoverybench}, and ScienceAgentBench~\cite{chen2025scienceagentbench}. While these approaches substantially reduce the manual effort required for scientific reproduction, their objective is fundamentally different from ours. Successful reproduction alone cannot determine whether an implementation faithfully matches the reported methodology, and implementation inspection alone cannot verify empirical claims. ReAgent bridges this gap by combining static implementation auditing with dynamic evidence collection to assess whether scientific claims are supported by their corresponding implementations and execution evidence.

\section{Method}

\begin{figure*}[t]
    \centering
     \includegraphics[width=0.9\textwidth]{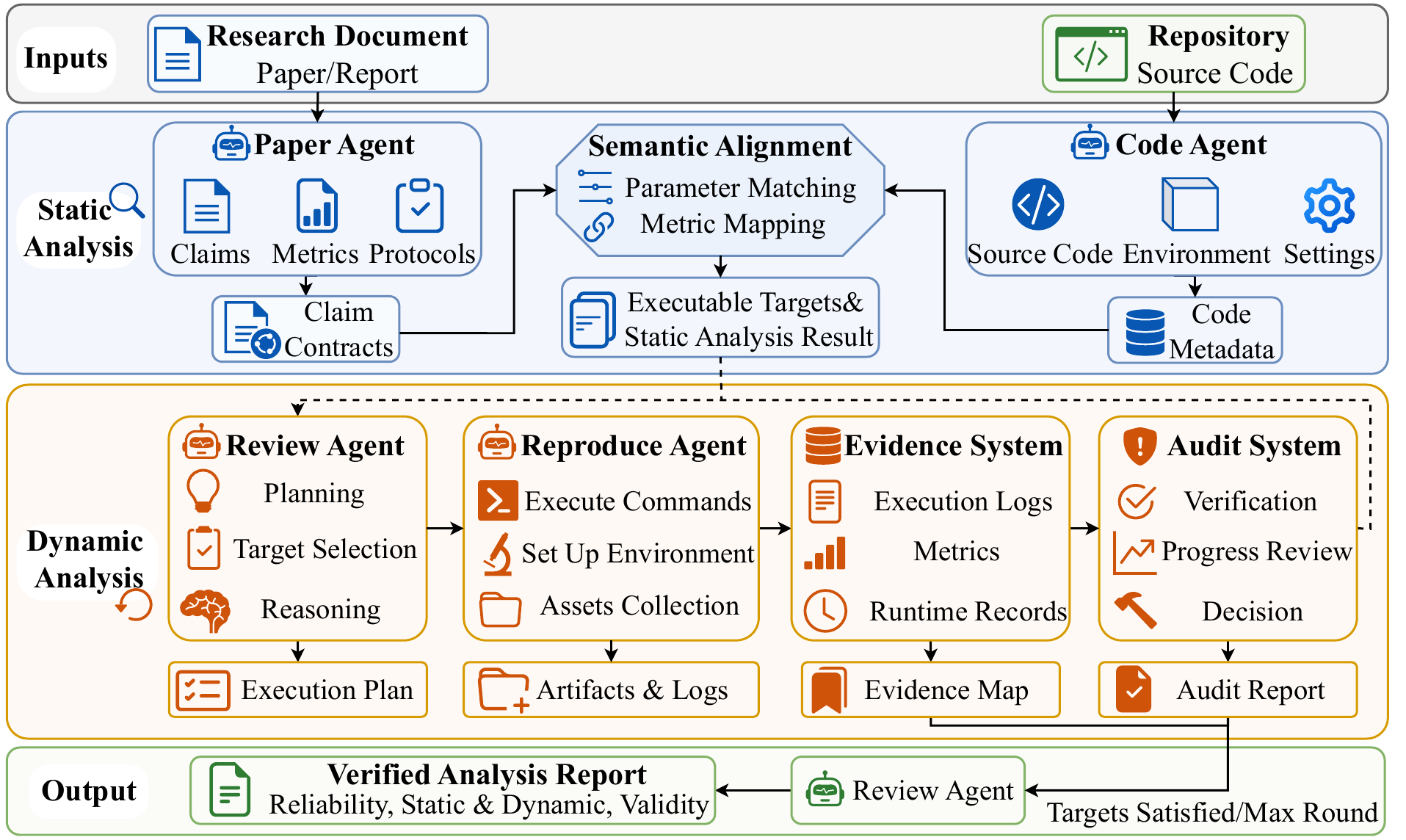}
    \caption{ReAgent framework.
Static verification constructs claim contracts, repository metadata, and semantic bindings to align scientific claims with implementation evidence and generate executable verification targets.
Dynamic verification performs review-guided reproduction and organizes execution feedback into provenance-aware evidence, enabling evidence-grounded audit decisions.}
    \label{fig:framework}
\end{figure*}

This section first introduces the concept of claim-code consistency auditing and then presents a two-stage verification framework for auditing automatically generated research documents and their associated repositories.
Given a scientific artifact containing research claims and its corresponding repository, ReAgent constructs structured claim contracts, repository metadata, and semantic bindings in the static stage, and transforms them into executable verification targets. 
In the dynamic stage, the Review Agent selects verification targets and plans verification strategies, while the Reproduce Agent executes the repository and collects execution evidence. 
The system then produces evidence-grounded consistency assessments and aggregates them into a verified analysis report.

\subsection{Problem Definition}
\label{sec:taxonomy}
Automated research agents are increasingly capable of conducting research and generating repositories. Existing studies primarily focus on the task completion rates of agentic research or the novelty of scientific ideas. However, these agents may intentionally or inadvertently produce inconsistent scientific outcomes. To study this challenge, we consider an automated research agent that independently generates both a research document and its accompanying repository. Our goal is not to model adversarial behavior, but to characterize the types of claim-code inconsistencies that may emerge during autonomous research generation, regardless of whether they are introduced intentionally or unintentionally.
Here, \(P\) denotes the textual research output containing scientific claims to be audited, including conventional papers as well as automatically generated research reports, while \(R\) denotes the associated repository containing implementation code and experimental configurations. An overview of this problem setting is illustrated in Figure~\ref{fig:Problem-Definition}.
The agent is regarded as a potential risk source, operating with the freedom to alter both research document text and repository during the research process. Consequently, it may produce seemingly plausible scientific artifacts that conceal underlying discrepancies between the research document and its implementation. Such inconsistencies can arise unintentionally—from planning errors, execution failures, incomplete implementations, or hallucinated experimental results—or intentionally, such as when an agent optimizes the narrative for persuasiveness without ensuring alignment with the repository.

Assuming the agent possesses unrestricted access to both the research document and the repository, we categorize the resulting inconsistencies into two major classes:
\begin{itemize}
\item \textbf{Static Repository Inconsistencies:} Discrepancies where the repository fails to align with the claims made in the research document  without requiring code execution.
\begin{itemize}
    \item Hardcoded Results: Reported figures are hardcoded into scripts or output templates rather than being generated through actual computation.
    \item Substantive Implementation Discrepancies: The reported module exists and runs, but its core algorithmic steps diverge substantially from the textual description.
    \item Omitted Implementation: Proposed methods or modules mentioned in the research document are absent from the repository.
    \item Shell/Dummy Implementation: Modules are defined, but they are short-circuited or never invoked during actual runtime.
    \item Fabricated Datasets: The research document claims to use a specific dataset, but the code generates synthetic data or uses a minor subset instead.
    \item Fabricated Experimental Hyperparameters: Claimed configurations or parameters are either omitted in code or fixed to arbitrary constant values.
    \item Data~/ Label Leakage: The test set, answers, or labels are exposed during training, hyperparameter tuning, or random seed selection.
\end{itemize}
\item \textbf{Dynamic Execution Inconsistencies:} Discrepancies between the actual execution outputs and the claims in the research document, typically involving exaggerated performance. Under identical configurations, the executed code yields performance far below the claimed results or fails to execute entirely.
\end{itemize}

Given a research document $P$ and its corresponding repository $R$, our objective is to determine whether scientific claims in $P$ are backed by verifiable evidence within $R$. To audit these claims, our framework, ReAgent, employs a complementary static-and-dynamic verification mechanism.

\subsection{Framework Overview}

Figure~\ref{fig:framework} illustrates the overall framework. The system takes a research document-repository pair \((P,R)\) as input and outputs an audit report \(A\). The framework consists of two stages. The first stage, Static Claim Verification, extracts structured information from both the research document and the repository, aligns scientific claims with their corresponding implementations, identifies potential static inconsistencies, and constructs executable specifications for subsequent dynamic verification.
The second stage, Dynamic Claim Verification, iteratively executes the repository under the guidance of a Review Agent. Runtime observations are organized into structured evidence associated with each claim, enabling dynamic verification based on execution outcomes.

\subsection{Static Claim Verification}

Static Claim Verification performs claim-guided analysis of the correspondence between scientific claims and repository implementations without executing code. It constructs structured claim representations, retrieves and aligns relevant implementation evidence, identifies potential inconsistencies, and establishes executable targets for subsequent dynamic verification. This phase comprises four steps: Claim Contract Construction, Coarse-to-Targeted Repository Analysis, Semantic Binding, and Executable Target Construction.

First, the system extracts scientific claims from the research document and transforms them into structured claim contracts. Each claim contract records the claim type, involved components, datasets, evaluation metrics, experimental settings, reported results, and their locations in the research document. These structured representations provide a unified interface for subsequent repository analysis and evidence alignment.

Next, the repository is analyzed progressively. The system first examines lightweight artifacts (e.g., README files, configuration files, and entry scripts) to obtain high-level execution information, and then focuses on artifacts-related code regions to inspect implementations, parameter interfaces, metric computation, and execution workflows.

The extracted information from the research document and repository is then aligned to associate each scientific claim with its corresponding implementation, including related modules, datasets, hyperparameters, and evaluation metrics. This alignment narrows the search space for dynamic verification while preserving explicit evidence provenance.

Based on the established correspondence, the Review Agent examines whether sufficient implementation evidence supports the reported claims and identifies potential issues such as missing implementations, parameter inconsistencies, or broken dependencies. For claims requiring execution-based validation, the system generates executable targets describing how each claim should be verified, including required resources, expected outputs, and evidence to collect.

\subsection{Dynamic Claim Verification}

Dynamic Claim Verification validates empirical claims by executing the repository and collecting execution evidence. Unlike static verification, which examines implementation consistency without execution, this stage evaluates whether experimental claims can be supported through actual execution. Starting from the executable targets generated in the static stage, the system iteratively plans, executes, and evaluates verification until sufficient evidence is collected or a predefined stopping condition is reached.

For each executable target derived from a claim, the Review Agent first generates a verification plan that defines the target claim, execution strategy, expected observations, and stopping criteria. Based on this plan, the Reproduce Agent prepares the execution environment, resolves software dependencies, acquires required datasets when available, and launches the corresponding experiments. To preserve the integrity of the audit, the agent is allowed to perform only minor compatibility fixes, such as correcting file paths or environment configurations, while modifications that alter the experimental methodology or reported algorithms are prohibited.

During execution, the evidence system continuously records runtime logs, command histories, intermediate artifacts, and generated outputs. These execution traces are automatically organized into structured evidence associated with the corresponding scientific claims. Besides numerical results, the collected evidence also includes contextual information such as execution commands, model configurations, datasets, timestamps, and artifact locations, ensuring that every verification result can be traced back to its original source.

Verification proceeds iteratively. After each execution round, the audit system evaluates the accumulated evidence to determine whether sufficient support has been collected for the current claim. If the verification criteria are satisfied, the framework terminates the execution phase and proceeds to audit report generation. Otherwise, the Review Agent analyzes the current evidence, identifies missing or conflicting information, and updates the execution plan by refining the search strategy, selecting alternative execution paths, or requesting additional experiments. Throughout the process, the system monitors execution progress to detect repeated failures, ineffective trials, and prolonged inactivity, preventing unnecessary exploration. The verification process terminates when all executable claims have been sufficiently verified or the maximum execution budget is exhausted. 

The output of this stage is a collection of execution-based verification results together with claim-level evidence, which are subsequently integrated with the static verification results during the final assessment.

\subsection{Final Assessment}

Finally, static verification results and execution evidence are jointly considered to produce the final audit assessment. Static verification identifies potential implementation-level inconsistencies, while dynamic verification provides execution-based evidence for empirical claims. For empirically verifiable claims, the system assigns dynamic verification outcomes, including \textit{Matched}, \textit{Mismatched}, \textit{Blocked}, or \textit{Not Reproduced}, together with supporting evidence. These findings are aggregated into a repository-level audit report, providing an overall assessment of artifact trustworthiness with complete evidence traceability.

\section{Experiments}

We evaluate whether ReAgent can reliably audit claim-code consistency in research artifacts generated by automated research agents. Our evaluation addresses three research questions:

\begin{itemize}
    \item \textbf{RQ1.} How accurately can ReAgent audit claim-code consistency compared with existing approaches?
    \item \textbf{RQ2.} What types of unsupported claims commonly appear in agent-generated research artifacts?
    \item \textbf{RQ3.} How does ReAgent perform on representative real-world auditing cases?
\end{itemize}

\subsection{Experimental Setup}
\label{sec:Expe}
Our evaluation consists of three components: the evaluation corpus, the claim auditing protocol for constructing ground truth, and comparison methods. Additional implementation details are provided in the supplementary material.

\subsubsection{Evaluation Corpus}

We construct our benchmark from AutoSOTA~\cite{li2026autosota}, which releases 188 repositories generated by automated research agents. Each repository contains an agent-generated report, source code, and experimental results. We assess the auditability of all 188 repositories and identify 92 repositories with insufficient evidence for reliable verification. These cases are treated as \emph{unverifiable} rather than inconsistent, leaving 96 auditable repositories for our analysis (Table~\ref{tab:dataset_filter}). Because this filtering may introduce selection bias, our reported inconsistency rates characterize the auditable subset rather than the full collection.

\begin{table}[t]
\centering
\small
\begin{tabular}{lcc}
\toprule
Reason & Description & Count\\
\midrule
Missing report & Report unavailable & 3\\
Missing repository & Source code unavailable & 10\\
Unavailable assets & Dataset/checkpoint unavailable & 79\\
\bottomrule
\end{tabular}
\caption{Repositories excluded during corpus construction.}
\label{tab:dataset_filter}
\end{table}

\subsubsection{Comparison Methods}

We compare ReAgent with two baselines.

\textbf{Luo et al.}~\cite{luo2025more} propose an auditing framework for automated scientific discovery that detects implementation issues such as data leakage, benchmark misuse, and methodological flaws. Since their framework does not explicitly verify claim-code consistency, we adapt it as a static verification baseline, referred to as Static-Audit.

\textbf{Codex}~\cite{openai2026codex} is a general-purpose coding agent with strong software engineering capabilities. It receives the same paper, repository, task description, and execution environment as ReAgent. 

All methods use GPT-5.4~\cite{openai2026gpt54} as the underlying model. Since different methods adopt different risk taxonomies, we evaluate them under a unified repository-level binary detection setting: a repository is positive if it contains at least one consistency issue.

\subsubsection{Evaluation Protocol}

We manually audit all repositories to establish ground truth.

For static verification, Claude Code is used only for candidate issue discovery. Three Claude models independently analyze repositories, and three authors verify all identified issues based on implementation details and experimental configurations. Disagreements are resolved through discussion, and repositories without detected issues are additionally inspected to reduce false negatives.

For dynamic verification, agents reproduce experiments in isolated environments. Execution traces are manually examined to verify whether reported results are supported by actual execution. Failed reproductions are further analyzed, with manual reproduction performed when necessary.

\begin{table*}[t]
\centering
\begin{tabular}{lcccccc}
\toprule
Method &
\multicolumn{3}{c}{Static Verification} &
Reproduction &
Dynamic Verification &
Overall Audit\\
\cmidrule(lr){2-4}
&
Accuracy~(\%) &
Recall~(\%) &
F1~(\%) &
Success Rate~(\%) &
 Accuracy~(\%) &
Accuracy~(\%)\\
\midrule

Static-Audit
& 31.25 & 100.00 & 45.90 & -- & -- & -- \\
Codex 
& 41.67 & 67.86 & 40.43 & 60.42 & 79.17 & 35.42 \\
ReAgent
& 78.13 & 82.14 & 68.66 & 76.04 & 90.63 & 70.83 \\

\bottomrule
\end{tabular}
\caption{Comparison of auditing performance.}
\label{tab:rq1}
\end{table*}

\subsection{RQ1: How accurately can ReAgent audit claim-code consistency?}

RQ1 evaluates whether ReAgent can correctly audit agent-generated repositories.
We compare the predicted auditing results against the manually established ground truth described in Section~\ref{sec:Expe}.

\paragraph{Metrics.}

\textbf{Static Verification Accuracy}
measures the proportion of repositories whose static verification decisions agree with manual annotations.
The static verification task is formulated as a binary classification problem, where the system predicts whether a repository contains static claim-code inconsistencies.
We additionally report Recall and F1-score due to the imbalanced distribution of inconsistent repositories.

\textbf{Reproduction Success Rate}
measures the proportion of repositories for which the auditing system successfully establishes an executable environment and completes experiment execution.

\textbf{Dynamic Verification Accuracy}
measures the proportion of repositories for which the binary dynamic verification decisions agree with the manually established ground truth.
Since our objective is to determine whether a scientific claim is supported by executable evidence, only Matched is regarded as a positive outcome. Mismatched, Blocked, and Not Reproduced are conservatively mapped to the negative class because none of them provides sufficient evidence to support the reported claim. Nevertheless, these outcomes are preserved as diagnostic labels in the audit report to differentiate distinct causes of verification failure.

\textbf{Overall Audit Accuracy}
measures the proportion of repositories for which both static verification and dynamic verification decisions agree with manual annotations.
The metric is computed over all evaluated repositories.

As shown in Table~\ref{tab:rq1}, ReAgent consistently outperforms existing methods across all evaluation dimensions. 
For static verification, ReAgent achieves 78.13\% accuracy, 82.14\% recall, and 68.66\% F1-score, substantially improving over Codex and Static-Audit. 
In particular, compared with Static-Audit, which achieves high recall but a low F1-score, ReAgent provides a better balance between identifying inconsistent repositories and avoiding false alarms.
This demonstrates the effectiveness of our structured claim--repository alignment for static consistency checking.

For dynamic verification, ReAgent achieves an accuracy of 90.63\% over all evaluated repositories, outperforming Codex by 11.46 percentage points. 
The improvement mainly comes from our execution-oriented verification pipeline, which converts empirical claims into executable verification targets and collects execution evidence through an integrated auditing and evidence management process.
By grounding verification decisions on experimental evidence rather than textual reasoning alone, ReAgent reduces unreliable judgments caused by incomplete or hallucinated evidence.

We further analyze dynamic verification failures to understand the limitations of execution-based auditing. Among failed cases, most failures were caused by practical execution barriers, including broken evaluation pipelines, missing datasets or assets, and incomplete repositories, which prevented the system from obtaining reliable execution evidence. 

Overall, ReAgent achieves 70.83\% audit accuracy, improving over Codex by 35.41 percentage points.
These results demonstrate that ReAgent enables a more reliable end-to-end consistency audit of paper--code artifacts generated by autonomous research agents.

\subsection{RQ2: What consistency issues are prevalent in agent-generated research artifacts?}

We next analyze what types of unsupported scientific claims are commonly produced by autonomous research agents. Unlike RQ1, which evaluates auditing accuracy, this analysis focuses on the characteristics of the detected inconsistencies.

Among the 96 repositories, 28 contain at least one static consistency issue. Since one repository may contain multiple issues, category counts do not sum to the number of affected repositories. Table~\ref{tab:risk_stat} summarizes the distribution. The most common issues are Substantive Implementation Discrepancies (10) and Data~/ Label Leakage (7), suggesting that autonomous research agents tend to modify implementations for performance improvement while failing to faithfully document these changes.

Besides these major categories, we also observe several less frequent but practically important inconsistency patterns, summarized as Others in Table~\ref{tab:risk_stat}. These cases mainly involve documentation and reproducibility risks, such as missing dependencies, undocumented configuration choices, or credential exposure. Although these issues do not fall into our seven predefined inconsistency categories, they can nevertheless impair the reproducibility of the repository, making it difficult for independent researchers to reliably reproduce or validate the reported findings.

Dynamic auditing further shows that 33 repositories successfully reproduce their reported metrics, while only 25 repositories satisfy both static consistency and successful reproduction. The remaining eight repositories reveal an important failure mode: their reported numerical results can be reproduced, while their implementations contain unsupported claim-code inconsistencies. Dynamic verification alone incorrectly considers these repositories consistent because the reported metrics are recovered. In contrast, the static verification stage of ReAgent detects seven of these eight cases, demonstrating that execution-based verification alone is insufficient for assessing the faithfulness of scientific claims.

These results demonstrate that neither static nor dynamic auditing alone is sufficient. Static verification is effective for identifying unsupported implementations and fabricated experimental settings, whereas dynamic verification is necessary to validate empirical claims through actual execution. Combining both provides substantially broader coverage of claim-code inconsistencies.

\begin{table}[t]
\centering
\small

\begin{tabular}{cc}
\toprule
Issue Type  & Count  \\
\midrule
Substantive Implementation Discrepancies & 10 \\
Data~/ Label Leakage & 7 \\
Fabricated Experimental Hyperparameters & 4  \\
Omitted Implementation& 4 \\
Shell/Dummy Implementation & 1  \\
Hardcoded Results  & 1  \\
Fabricated Datasets & 0  \\
Others & 6  \\
\bottomrule
\end{tabular}
\caption{Distribution of static consistency issues identified by manual auditing.}
\label{tab:risk_stat}
\end{table}

\subsection{RQ3: How does ReAgent perform on representative real-world auditing cases?}

\subsubsection{Static verification identifies implementation inconsistencies}

One repository claims that propensity-feature augmentation with robust ensembling improves IHDP PEHE from 0.1829 to 0.1539. Static verification revealed that the evaluation pipeline implemented an S-learner augmentation strategy rather than the claimed propensity-feature augmentation, indicating a substantive implementation discrepancy. This case shows that execution alone may miss inconsistencies when the repository remains runnable but deviates from the reported methodology.

\subsubsection{Dynamic verification avoids false negatives}
Another repository reports that Anchorage AUROC improves from 99.31 to 99.42 after twelve optimization iterations. ReAgent reproduced the official evaluation pipeline and obtained the reported value of 99.42. In contrast, Codex used a simplified reproduction strategy due to computational constraints and incorrectly concluded that the improvement was unverifiable, highlighting the importance of faithful execution for reliable auditing.
\subsubsection{Dynamic verification distinguishes blocked reproduction}
Beyond binary claim-code consistency classification, dynamic verification further distinguishes genuine claim-code inconsistencies from cases where reproduction is blocked by missing or unusable execution artifacts.

Another repository reports a PSNR of 32.24 obtained from its official seven-seed optimization pipeline. Static verification confirmed that the documented workflow and implementation were consistent with the reported methodology. However, during dynamic verification, ReAgent could only execute the evaluation stage, reproducing a PSNR of 30.66 instead of the reported 32.24. Further investigation revealed that the complete generation pipeline failed because the required model checkpoint (\texttt{weights/PFCNet.pth}) was an invalid HTML page rather than a valid PyTorch checkpoint. ReAgent therefore classified this case as \textit{Blocked} rather than \textit{Mismatched}, distinguishing missing executable artifacts from genuine claim-code inconsistencies.

\section{Discussion and Limitations}
Our study suggests that claim-code inconsistency represents an emerging challenge for autonomous research agents. As agents iteratively revise manuscripts, code, and experimental configurations, inconsistencies may arise even without intentional fabrication, potentially reducing the trustworthiness and reproducibility of generated scientific artifacts.
Despite the encouraging results, our framework has several limitations. 

First, ReAgent evaluates whether scientific claims are supported by the accompanying repository, rather than assessing the scientific validity or novelty of the proposed methods. Therefore, a repository may be internally consistent while still containing flawed assumptions, inappropriate experimental designs, or incorrect scientific conclusions beyond the scope of claim-code consistency.
Second, dynamic verification relies on executable research artifacts. Missing datasets, unavailable checkpoints, obsolete dependencies, or excessive computational requirements may prevent complete reproduction even when implementations are correct. Although ReAgent provides isolated execution environments and automated dependency handling, these issues remain challenging for large-scale auditing.

Finally, our evaluation is conducted exclusively on AutoSOTA, currently the only autonomous research platform we identified that publicly releases paper--repository pairs at scale. Therefore, the benchmark may contain distribution and survivorship biases, and the results may not fully generalize to other agents or human researchers. Nevertheless, using a single source reduces generator-specific variation and enables a controlled evaluation of claim-code consistency.

\section{Conclusion}

This paper presents ReAgent, an automated auditing framework for assessing the consistency between agent-generated research documents and their associated repositories. By combining claim-guided static repository verification with execution-based dynamic verification, ReAgent systematically collects and organizes implementation and execution evidence to support repository-level audit decisions. Experiments demonstrate that ReAgent effectively identifies claim-code inconsistencies while providing transparent and traceable auditing evidence.

We believe this work highlights claim-code consistency as a fundamental challenge for trustworthy autonomous research. As research agents increasingly generate papers, code, and experiments end-to-end, assessing scientific capability alone is insufficient; reliable autonomous research also requires verifying that scientific claims remain faithfully grounded in their implementations and execution evidence. By introducing a systematic auditing framework for this emerging setting, ReAgent provides a step toward more transparent and reliable evaluation of autonomous scientific workflows. Future work will extend ReAgent to finer-grained claim-level auditing, broader scientific domains, richer evidence sources, and larger-scale auditing benchmarks.

\bibliography{aaai2027}

@article{wang2024survey,
  title={A survey on large language model based autonomous agents},
  author={Wang, Lei and Ma, Chen and Feng, Xueyang and Zhang, Zeyu and Yang, Hao and Zhang, Jingsen and Chen, Zhiyuan and Tang, Jiakai and Chen, Xu and Lin, Yankai and others},
  journal={Frontiers of Computer Science},
  volume={18},
  number={6},
  pages={186345},
  year={2024},
  publisher={Springer}
}

@inproceedings{jia2026autotool,
  author       = {Jingyi Jia and
                  Qinbin Li},
  editor       = {Sven Koenig and
                  Chad Jenkins and
                  Matthew E. Taylor},
  title        = {AutoTool: Efficient Tool Selection for Large Language Model Agents},
  booktitle    = {Fortieth {AAAI} Conference on Artificial Intelligence, Thirty-Eighth
                  Conference on Innovative Applications of Artificial Intelligence,
                  Sixteenth Symposium on Educational Advances in Artificial Intelligence,
                  {AAAI} 2026, Singapore, January 20-27, 2026},
  pages        = {31265--31273},
  publisher    = {{AAAI} Press},
  year         = {2026},
  url          = {https://doi.org/10.1609/aaai.v40i37.40389},
  doi          = {10.1609/AAAI.V40I37.40389},
  bibsource    = {dblp computer science bibliography, https://dblp.org}
}

@inproceedings{jimenez2024swe,
  author       = {Carlos E. Jimenez and
                  John Yang and
                  Alexander Wettig and
                  Shunyu Yao and
                  Kexin Pei and
                  Ofir Press and
                  Karthik R. Narasimhan},
  title        = {SWE-bench: Can Language Models Resolve Real-world Github Issues?},
  booktitle    = {The Twelfth International Conference on Learning Representations,
                  {ICLR} 2024, Vienna, Austria, May 7-11, 2024},
  publisher    = {OpenReview.net},
  year         = {2024},
  url          = {https://openreview.net/forum?id=VTF8yNQM66},
  bibsource    = {dblp computer science bibliography, https://dblp.org}
}

@inproceedings{yang2024swe,
  author       = {John Yang and
                  Carlos E. Jimenez and
                  Alexander Wettig and
                  Kilian Lieret and
                  Shunyu Yao and
                  Karthik Narasimhan and
                  Ofir Press},
  editor       = {Amir Globersons and
                  Lester Mackey and
                  Danielle Belgrave and
                  Angela Fan and
                  Ulrich Paquet and
                  Jakub M. Tomczak and
                  Cheng Zhang},
  title        = {SWE-agent: Agent-Computer Interfaces Enable Automated Software Engineering},
  booktitle    = {Advances in Neural Information Processing Systems 37: Annual Conference
                  on Neural Information Processing Systems 2024, NeurIPS 2024, Vancouver,
                  BC, Canada, December 10 - 15, 2024},
  year         = {2024},
  url          = {http://papers.nips.cc/paper\_files/paper/2024/hash/5a7c947568c1b1328ccc5230172e1e7c-Abstract-Conference.html},
  bibsource    = {dblp computer science bibliography, https://dblp.org}
}

@inproceedings{liu2025open,
  author       = {Xiaotian Liu and
                  Ali Pesaranghader and
                  Hanze Li and
                  Punyaphat Sukcharoenchaikul and
                  Jaehong Kim and
                  Tanmana Sadhu and
                  Hyejeong Jeon and
                  Scott Sanner},
  editor       = {Wanxiang Che and
                  Joyce Nabende and
                  Ekaterina Shutova and
                  Mohammad Taher Pilehvar},
  title        = {Open-World Planning via Lifted Regression with LLM-Inferred Affordances
                  for Embodied Agents},
  booktitle    = {Proceedings of the 63rd Annual Meeting of the Association for Computational
                  Linguistics (Volume 1: Long Papers), {ACL} 2025, Vienna, Austria,
                  July 27 - August 1, 2025},
  pages        = {20881--20897},
  publisher    = {Association for Computational Linguistics},
  year         = {2025},
  url          = {https://doi.org/10.18653/v1/2025.acl-long.1018},
  doi          = {10.18653/V1/2025.ACL-LONG.1018},
  bibsource    = {dblp computer science bibliography, https://dblp.org}
}

@inproceedings{lin2023grounded,
  author       = {Bill Yuchen Lin and
                  Chengsong Huang and
                  Qian Liu and
                  Wenda Gu and
                  Sam Sommerer and
                  Xiang Ren},
  editor       = {Brian Williams and
                  Yiling Chen and
                  Jennifer Neville},
  title        = {On Grounded Planning for Embodied Tasks with Language Models},
  booktitle    = {Thirty-Seventh {AAAI} Conference on Artificial Intelligence, {AAAI}
                  2023, Thirty-Fifth Conference on Innovative Applications of Artificial
                  Intelligence, {IAAI} 2023, Thirteenth Symposium on Educational Advances
                  in Artificial Intelligence, {EAAI} 2023, Washington, DC, USA, February
                  7-14, 2023},
  pages        = {13192--13200},
  publisher    = {{AAAI} Press},
  year         = {2023},
  url          = {https://doi.org/10.1609/aaai.v37i11.26549},
  doi          = {10.1609/AAAI.V37I11.26549},
  bibsource    = {dblp computer science bibliography, https://dblp.org}
}

@inproceedings{zheng2025automation,
  author       = {Tianshi Zheng and
                  Zheye Deng and
                  Hong Ting Tsang and
                  Weiqi Wang and
                  Jiaxin Bai and
                  Zihao Wang and
                  Yangqiu Song},
  editor       = {Christos Christodoulopoulos and
                  Tanmoy Chakraborty and
                  Carolyn Rose and
                  Violet Peng},
  title        = {From Automation to Autonomy: {A} Survey on Large Language Models in
                  Scientific Discovery},
  booktitle    = {Proceedings of the 2025 Conference on Empirical Methods in Natural
                  Language Processing, {EMNLP} 2025, Suzhou, China, November 4-9, 2025},
  pages        = {17733--17750},
  publisher    = {Association for Computational Linguistics},
  year         = {2025},
  url          = {https://doi.org/10.18653/v1/2025.emnlp-main.895},
  doi          = {10.18653/V1/2025.EMNLP-MAIN.895},
  bibsource    = {dblp computer science bibliography, https://dblp.org}
}

@inproceedings{zamprogno2025autonomous,
  author       = {Giacomo Zamprogno and
                  Ilaria Tiddi and
                  Bart Verheij},
  editor       = {Ron P. A. Petrick and
                  Christopher W. Geib},
  title        = {Autonomous Research Assistants for Hybrid Intelligence: Landscape
                  and Challenges},
  booktitle    = {Proceedings of the 2025 {AAAI} Spring Symposium Series, San Francisco,
                  CA, USA, March 31-April 2, 2025},
  pages        = {350--358},
  publisher    = {{AAAI} Press},
  year         = {2025},
  url          = {https://doi.org/10.1609/aaaiss.v5i1.35611},
  doi          = {10.1609/AAAISS.V5I1.35611},
  bibsource    = {dblp computer science bibliography, https://dblp.org}
}

@inproceedings{su2025many,
  author       = {Haoyang Su and
                  Renqi Chen and
                  Shixiang Tang and
                  Zhenfei Yin and
                  Xinzhe Zheng and
                  Jinzhe Li and
                  Biqing Qi and
                  Qi Wu and
                  Hui Li and
                  Wanli Ouyang and
                  Philip Torr and
                  Bowen Zhou and
                  Nanqing Dong},
  editor       = {Wanxiang Che and
                  Joyce Nabende and
                  Ekaterina Shutova and
                  Mohammad Taher Pilehvar},
  title        = {Many Heads Are Better Than One: Improved Scientific Idea Generation
                  by {A} LLM-Based Multi-Agent System},
  booktitle    = {Proceedings of the 63rd Annual Meeting of the Association for Computational
                  Linguistics (Volume 1: Long Papers), {ACL} 2025, Vienna, Austria,
                  July 27 - August 1, 2025},
  pages        = {28201--28240},
  publisher    = {Association for Computational Linguistics},
  year         = {2025},
  url          = {https://doi.org/10.18653/v1/2025.acl-long.1368},
  doi          = {10.18653/V1/2025.ACL-LONG.1368},
  bibsource    = {dblp computer science bibliography, https://dblp.org}
}

@article{lu2026towards,
  title={Towards end-to-end automation of AI research},
  author={Lu, Chris and Lu, Cong and Lange, Robert Tjarko and Yamada, Yutaro and Hu, Shengran and Foerster, Jakob and Ha, David and Clune, Jeff},
  journal={Nature},
  volume={651},
  number={8107},
  pages={914},
  year={2026}
}

@inproceedings{schmidgall2025agent,
  author       = {Samuel Schmidgall and
                  Yusheng Su and
                  Ze Wang and
                  Ximeng Sun and
                  Jialian Wu and
                  Xiaodong Yu and
                  Jiang Liu and
                  Michael Moor and
                  Zicheng Liu and
                  Emad Barsoum},
  editor       = {Christos Christodoulopoulos and
                  Tanmoy Chakraborty and
                  Carolyn Rose and
                  Violet Peng},
  title        = {Agent Laboratory: Using {LLM} Agents as Research Assistants},
  booktitle    = {Findings of the Association for Computational Linguistics: {EMNLP}
                  2025, Suzhou, China, November 4-9, 2025},
  pages        = {5977--6043},
  publisher    = {Association for Computational Linguistics},
  year         = {2025},
  url          = {https://doi.org/10.18653/v1/2025.findings-emnlp.320},
  doi          = {10.18653/V1/2025.FINDINGS-EMNLP.320},
  bibsource    = {dblp computer science bibliography, https://dblp.org}
}

@misc{tang2026fars,
      title={FARS: A Fully Automated Research System Deployed at Scale}, 
      author={Qiong Tang and Tianxiang Sun and Xiangkun Hu and Xiangyang Liu and Yiran Chen and Yunfan Shao and Bobo Li and Changze Lv and Cheng Xu and Chengsong Huang and Chunyang Li and Dizhan Xue and Hao Bai and Haodong Duan and Hengquan Guo and Hongyang He and Hongyi Chen and Hui Shen and Jiahao Yuan and Jiankai Sun and Jikang Cheng and Jinfeng Xu and Jingqi Tong and Jingye Chen and Jinxiu Liu and Jixuan Leng and Junchi Yu and Kaixun Jiang and Kun Xiang and Kunpeng Yao and Lang Feng and Liangqi Yuan and Longsen Gao and Meng Li and Qi Jia and Qiushi Sun and Shengyuan Ding and Shizhan Gong and Siru Zhong and Terry Jingchen Zhang and Tianle Gu and Tianyi Liang and Weijie Liu and Weikai Yang and Weizhi Fei and Xin Wang and Xinpeng Liu and Xuanwen Ding and Yihong Tang and Yuanli Wang and Yukun Jiang and Yuming Yang and Zhengbao He and Zhikai Chen and Zhikun Xu and Zhuang Li and Zihao Huang},
      year={2026},
      eprint={2606.31651},
      archivePrefix={arXiv},
      primaryClass={cs.AI},
      url={https://arxiv.org/abs/2606.31651}, 
}

@misc{si2024can,
      title={Can LLMs Generate Novel Research Ideas? A Large-Scale Human Study with 100+ NLP Researchers}, 
      author={Chenglei Si and Diyi Yang and Tatsunori Hashimoto},
      year={2024},
      eprint={2409.04109},
      archivePrefix={arXiv},
      primaryClass={cs.CL},
      url={https://arxiv.org/abs/2409.04109}, 
}

@misc{chen2025mlrbench,
      title={MLR-Bench: Evaluating AI Agents on Open-Ended Machine Learning Research}, 
      author={Hui Chen and Miao Xiong and Yujie Lu and Wei Han and Ailin Deng and Yufei He and Jiaying Wu and Yibo Li and Yue Liu and Bryan Hooi},
      year={2025},
      eprint={2505.19955},
      archivePrefix={arXiv},
      primaryClass={cs.LG},
      url={https://arxiv.org/abs/2505.19955}, 
}

@misc{wang2026act,
      title={Act As a Real Researcher: A Suite of Benchmarks Evaluating Frontier LLMs and Agentic Harnesses in Research Lifecycle}, 
      author={Jiayu Wang and Weijiang Lv and Bowen Fu and Jing Fu and Jiayi Song and Lingyu Zhang and Lanxuan Xue and Luodi Chen and Zepeng Xin and Kaiyu Li and Xiangyong Cao},
      year={2026},
      eprint={2606.07462},
      archivePrefix={arXiv},
      primaryClass={cs.AI},
      url={https://arxiv.org/abs/2606.07462}, 
}

@misc{baker20161,
  title={1,500 scientists lift the lid on reproducibility},
  author={Baker, Monya},
  year={2016},
  publisher={Nature Publishing Group UK London}
}

@inproceedings{gundersen2018state,
  author       = {Odd Erik Gundersen and
                  Sigbj{\o}rn Kjensmo},
  editor       = {Sheila A. McIlraith and
                  Kilian Q. Weinberger},
  title        = {State of the Art: Reproducibility in Artificial Intelligence},
  booktitle    = {Proceedings of the Thirty-Second {AAAI} Conference on Artificial Intelligence,
                  (AAAI-18), the 30th innovative Applications of Artificial Intelligence
                  (IAAI-18), and the 8th {AAAI} Symposium on Educational Advances in
                  Artificial Intelligence (EAAI-18), New Orleans, Louisiana, USA, February
                  2-7, 2018},
  
  publisher    = {{AAAI} Press},
  year         = {2018},
  url          = {https://doi.org/10.1609/aaai.v32i1.11503},
  doi          = {10.1609/AAAI.V32I1.11503},
pages        = {1644--1651},
  bibsource    = {dblp computer science bibliography, https://dblp.org}
}

@article{pineau2021improving,
  title={Improving reproducibility in machine learning research (a report from the neurips 2019 reproducibility program)},
  author={Pineau, Joelle and Vincent-Lamarre, Philippe and Sinha, Koustuv and Larivi{\`e}re, Vincent and Beygelzimer, Alina and d'Alch{\'e}-Buc, Florence and Fox, Emily and Larochelle, Hugo},
  journal={Journal of machine learning research},
  volume={22},
  number={164},
  pages={1--20},
  year={2021}
}

@inproceedings{wadden2020fact,
  title={Fact or fiction: Verifying scientific claims},
  author={Wadden, David and Lin, Shanchuan and Lo, Kyle and Wang, Lucy Lu and van Zuylen, Madeleine and Cohan, Arman and Hajishirzi, Hannaneh},
  booktitle={Proceedings of the 2020 Conference on Empirical Methods in Natural Language Processing (EMNLP)},
  pages={7534--7550},
  year={2020}
}

@misc{starace2025paperbench,
      title={PaperBench: Evaluating AI's Ability to Replicate AI Research}, 
      author={Giulio Starace and Oliver Jaffe and Dane Sherburn and James Aung and Jun Shern Chan and Leon Maksin and Rachel Dias and Evan Mays and Benjamin Kinsella and Wyatt Thompson and Johannes Heidecke and Amelia Glaese and Tejal Patwardhan},
      year={2025},
      eprint={2504.01848},
      archivePrefix={arXiv},
      primaryClass={cs.AI},
      url={https://arxiv.org/abs/2504.01848}, 
}

@inproceedings{chen2025deep,
  title={Deep-Reproducer: From Paper Understanding to Code Generation},
  author={Chen, Pengcheng and Yan, Ning and Zhao, Zihan and Lin, Yixiao and Chen, Huaibo and Hu, Yue and Bai, Qinbo and Li, Xiang and Mortazavi, Masood S},
    year={2025},
  booktitle={NeurIPS 2025 Fourth Workshop on Deep Learning for Code}
}

@misc{luo2025more,
      title={The More You Automate, the Less You See: Hidden Pitfalls of AI Scientist Systems}, 
      author={Ziming Luo and Atoosa Kasirzadeh and Nihar B. Shah},
      year={2025},
      eprint={2509.08713},
      archivePrefix={arXiv},
      primaryClass={cs.AI},
      url={https://arxiv.org/abs/2509.08713}, 
}

@inproceedings{chen2025scienceagentbench,
  author       = {Ziru Chen and
                  Shijie Chen and
                  Yuting Ning and
                  Qianheng Zhang and
                  Boshi Wang and
                  Botao Yu and
                  Yifei Li and
                  Zeyi Liao and
                  Chen Wei and
                  Zitong Lu and
                  Vishal Dey and
                  Mingyi Xue and
                  Frazier N. Baker and
                  Benjamin Burns and
                  Daniel Adu{-}Ampratwum and
                  Xuhui Huang and
                  Xia Ning and
                  Song Gao and
                  Yu Su and
                  Huan Sun},
  title        = {ScienceAgentBench: Toward Rigorous Assessment of Language Agents for
                  Data-Driven Scientific Discovery},
  booktitle    = {The Thirteenth International Conference on Learning Representations,
                  {ICLR} 2025, Singapore, April 24-28, 2025},
  publisher    = {OpenReview.net},
  year         = {2025},
  url          = {https://openreview.net/forum?id=6z4YKr0GK6},
  bibsource    = {dblp computer science bibliography, https://dblp.org}
}

@inproceedings{tang2026ai,
 author = {Tang, Jiabin and Xia, Lianghao and Li, Zhonghang and Huang, Chao},
 booktitle = {Advances in Neural Information Processing Systems},
 editor = {D. Belgrave and C. Zhang and H. Lin and R. Pascanu and P. Koniusz and M. Ghassemi and N. Chen},
 pages = {9481--9520},
 publisher = {Curran Associates, Inc.},
 title = {AI-Researcher: Autonomous Scientific Innovation},
 url = {https://proceedings.neurips.cc/paper_files/paper/2025/file/0d904d300a105809a2114d727851e759-Paper-Conference.pdf},
 volume = {38},
 year = {2025}
}

@misc{li2026autosota,
      title={AutoSOTA: An End-to-End Automated Research System for State-of-the-Art AI Model Discovery}, 
      author={Yu Li and Chenyang Shao and Xinyang Liu and Ruotong Zhao and Peijie Liu and Hongyuan Su and Zhibin Chen and Qinglong Yang and Anjie Xu and Yi Fang and Qingbin Zeng and Tianxing Li and Jingbo Xu and Fengli Xu and Yong Li and Tie-Yan Liu},
      year={2026},
      eprint={2604.05550},
      archivePrefix={arXiv},
      primaryClass={cs.CL},
      url={https://arxiv.org/abs/2604.05550}, 
}

@misc{meng2026scientistone,
      title={ScientistOne: Towards Human-Level Autonomous Research via Chain-of-Evidence}, 
      author={Rui Meng and Bhavana Dalvi Mishra and Jiefeng Chen and Chun-Liang Li and Palash Goyal and Mihir Parmar and Yiwen Song and Yale Song and Rajarishi Sinha and Parthasarathy Ranganathan and Burak Gokturk and Jinsung Yoon and Tomas Pfister},
      year={2026},
      eprint={2605.26340},
      archivePrefix={arXiv},
      primaryClass={cs.AI},
      url={https://arxiv.org/abs/2605.26340}, 
}

@misc{liu2026autoresearchclaw,
      title={AutoResearchClaw: Self-Reinforcing Autonomous Research with Human-AI Collaboration}, 
      author={Jiaqi Liu and Shi Qiu and Mairui Li and Bingzhou Li and Haonian Ji and Siwei Han and Xinyu Ye and Peng Xia and Zihan Dong and Meng Chen and Congyu Zhang and Letian Zhang and Guiming Chen and Haoqin Tu and Xinyu Yang and Lu Feng and Xujiang Zhao and Haifeng Chen and Jiawei Zhou and Xiao Wang and Weitong Zhang and Hongtu Zhu and Yun Li and Jieru Mei and Hongliang Fei and Jiaheng Zhang and Linjie Li and Linjun Zhang and Yuyin Zhou and Sheng Wang and Caiming Xiong and James Zou and Zeyu Zheng and Cihang Xie and Mingyu Ding and Huaxiu Yao},
      year={2026},
      eprint={2605.20025},
      archivePrefix={arXiv},
      primaryClass={cs.AI},
      url={https://arxiv.org/abs/2605.20025}, 
}

@inproceedings{huang2026deceptionbench,
 author = {Huang, Yao and Sun, Yitong and Zhang, Yichi and Zhang, Ruochen and Dong, Yinpeng and Wei, Xingxing},
 booktitle = {Advances in Neural Information Processing Systems},
 editor = {D. Belgrave and C. Zhang and H. Lin and R. Pascanu and P. Koniusz and M. Ghassemi and N. Chen},
 pages = {},
 publisher = {Curran Associates, Inc.},
 title = {DeceptionBench: A Comprehensive Benchmark for AI Deception Behaviors in Real-world Scenarios},
 url = {https://proceedings.neurips.cc/paper_files/paper/2025/file/55494d8756b72c2219027edc9de1ee5a-Paper-Datasets_and_Benchmarks_Track.pdf},
 volume = {38},
 year = {2025}
}

@inproceedings{DBLP:conf/ijcnlp/MathewMMVWCS25,
  author       = {Yohan Mathew and
                  Ollie Matthews and
                  Robert McCarthy and
                  Joan Velja and
                  Christian Schr{\"{o}}der de Witt and
                  Dylan Cope and
                  Nandi Schoots},
  editor       = {Kentaro Inui and
                  Sakriani Sakti and
                  Haofen Wang and
                  Derek F. Wong and
                  Pushpak Bhattacharyya and
                  Biplab Banerjee and
                  Asif Ekbal and
                  Tanmoy Chakraborty and
                  Dhirendra Pratap Singh},
  title        = {Hidden in Plain Text: Emergence {\&} Mitigation of Steganographic
                  Collusion in LLMs},
  booktitle    = {Proceedings of the 14th International Joint Conference on Natural
                  Language Processing and the 4th Conference of the Asia-Pacific Chapter
                  of the Association for Computational Linguistics, {IJCNLP-AACL} 2025,
                  Mumbai, India, December 20-24, 2025},
  pages        = {585--624},
  publisher    = {The Asian Federation of Natural Language Processing and The Association
                  for Computational Linguistics},
  year         = {2025},
  url          = {https://doi.org/10.18653/v1/2025.ijcnlp-long.34},
  doi          = {10.18653/V1/2025.IJCNLP-LONG.34},
  bibsource    = {dblp computer science bibliography, https://dblp.org}
}

@misc{majumder2024discoverybench,
      title={DiscoveryBench: Towards Data-Driven Discovery with Large Language Models}, 
      author={Bodhisattwa Prasad Majumder and Harshit Surana and Dhruv Agarwal and Bhavana Dalvi Mishra and Abhijeetsingh Meena and Aryan Prakhar and Tirth Vora and Tushar Khot and Ashish Sabharwal and Peter Clark},
      year={2024},
      eprint={2407.01725},
      archivePrefix={arXiv},
      primaryClass={cs.CL},
      url={https://arxiv.org/abs/2407.01725}, 
}

@misc{openai2026gpt54,
  author       = {{OpenAI}},
  title        = {{GPT-5.4: Technical Report and Model Card}},
  year         = {2026},
  howpublished = {\url{https://openai.com/index/introducing-gpt-5-4/}},
  note         = {Accessed: 2026-07-29}
}

@misc{openai2026codex,
  author       = {{OpenAI}},
  title        = {{OpenAI Codex: The Next Era of Knowledge Work and Code Generation}},
  year         = {2026},
  howpublished = {\url{https://openai.com/index/codex-for-knowledge-work/}},
  note         = {Accessed: 2026-07-29}
}

@article{ji2023survey,
  author       = {Ziwei Ji and
                  Nayeon Lee and
                  Rita Frieske and
                  Tiezheng Yu and
                  Dan Su and
                  Yan Xu and
                  Etsuko Ishii and
                  Yejin Bang and
                  Andrea Madotto and
                  Pascale Fung},
  title        = {Survey of Hallucination in Natural Language Generation},
  journal      = {{ACM} Comput. Surv.},
  volume       = {55},
  number       = {12},
  pages        = {248:1--248:38},
  year         = {2023},
  url          = {https://doi.org/10.1145/3571730},
  doi          = {10.1145/3571730},
  bibsource    = {dblp computer science bibliography, https://dblp.org}
}

@inproceedings{huang2024selfcorrect,
  author       = {Jie Huang and
                  Xinyun Chen and
                  Swaroop Mishra and
                  Huaixiu Steven Zheng and
                  Adams Wei Yu and
                  Xinying Song and
                  Denny Zhou},
  title        = {Large Language Models Cannot Self-Correct Reasoning Yet},
  booktitle    = {The Twelfth International Conference on Learning Representations,
                  {ICLR} 2024, Vienna, Austria, May 7-11, 2024},
  publisher    = {OpenReview.net},
  year         = {2024},
  url          = {https://openreview.net/forum?id=IkmD3fKBPQ},
  bibsource    = {dblp computer science bibliography, https://dblp.org}
}

@inproceedings{tyen2024llms,
  author       = {Gladys Tyen and
                  Hassan Mansoor and
                  Victor Carbune and
                  Peter Chen and
                  Tony Mak},
  editor       = {Lun{-}Wei Ku and
                  Andre Martins and
                  Vivek Srikumar},
  title        = {LLMs cannot find reasoning errors, but can correct them given the
                  error location},
  booktitle    = {Findings of the Association for Computational Linguistics, {ACL} 2024,
                  Bangkok, Thailand and virtual meeting, August 11-16, 2024},
  series       = {Findings of {ACL}},
  volume       = {{ACL} 2024},
  pages        = {13894--13908},
  publisher    = {Association for Computational Linguistics},
  year         = {2024},
  url          = {https://doi.org/10.18653/v1/2024.findings-acl.826},
  doi          = {10.18653/V1/2024.FINDINGS-ACL.826},
  bibsource    = {dblp computer science bibliography, https://dblp.org}
}

@inproceedings{tian2025codehalu,
  author       = {Yuchen Tian and
                  Weixiang Yan and
                  Qian Yang and
                  Xuandong Zhao and
                  Qian Chen and
                  Wen Wang and
                  Ziyang Luo and
                  Lei Ma and
                  Dawn Song},
  editor       = {Toby Walsh and
                  Julie Shah and
                  Zico Kolter},
  title        = {CodeHalu: Investigating Code Hallucinations in LLMs via Execution-based
                  Verification},
  booktitle    = {Thirty-Ninth {AAAI} Conference on Artificial Intelligence, Thirty-Seventh
                  Conference on Innovative Applications of Artificial Intelligence,
                  Fifteenth Symposium on Educational Advances in Artificial Intelligence,
                  {AAAI} 2025, Philadelphia, PA, USA, February 25 - March 4, 2025},
  pages        = {25300--25308},
  publisher    = {{AAAI} Press},
  year         = {2025},
  url          = {https://doi.org/10.1609/aaai.v39i24.34717},
  doi          = {10.1609/AAAI.V39I24.34717},
  bibsource    = {dblp computer science bibliography, https://dblp.org}
}

@inproceedings{baumgartner2026scicoqa,
  author       = {Tim Baumg{\"{a}}rtner and
                  Iryna Gurevych},
  editor       = {Maria Liakata and
                  Viviane P. Moreira and
                  Jiajun Zhang and
                  David Jurgens},
  title        = {SciCoQA: Quality Assurance for Scientific Paper-Code Alignment},
  booktitle    = {Proceedings of the 64th Annual Meeting of the Association for Computational
                  Linguistics (Volume 1: Long Papers), {ACL} 2026, San Diego, California,
                  United States, July 2-7, 2026},
  pages        = {38740--38770},
  publisher    = {Association for Computational Linguistics},
  year         = {2026},
  url          = {https://aclanthology.org/2026.acl-long.1795/},
  bibsource    = {dblp computer science bibliography, https://dblp.org}
}

\end{document}